# GPS Attack Detection and Mitigation for Safe Autonomous Driving using Image and Map based Lateral Direction Localization

Qingming Chen, Peng Liu, Guoqiang Li, Zhenpo Wang

*Abstract*—The accuracy and robustness of vehicle localization are critical for achieving safe and reliable high-level autonomy. Recent results show that GPS is vulnerable to spoofing attacks, which is one major threat to autonomous driving. In this paper, a novel anomaly detection and mitigation method against GPS attacks that utilizes onboard camera and high-precision maps is proposed to ensure accurate vehicle localization. First, lateral direction localization in driving lanes is calculated by camera-based lane detection and map matching respectively. Then, a real-time detector for GPS spoofing attack is developed to evaluate the localization data. When the attack is detected, a multi-source fusion-based localization method using Unscented Kalman filter is derived to mitigate GPS attack and improve the localization accuracy. The proposed method is validated in various scenarios in Carla simulator and open-source public dataset to demonstrate its effectiveness in timely GPS attack detection and data recovery.

## I. INTRODUCTION

In the past decades, there has been a significant improvement in the accuracy of positioning for autonomous vehicles. Through sensor fusion, autonomous vehicles are able to acquire information from multiple sources to determine their own position, thereby enhancing driving safety [1]. Accurate vehicle positioning is crucial for subsequent perception, planning, and control algorithms, in order to achieve the goal of executing correct driving decisions. GPS, as a mature global coordinate positioning system, plays an irreplaceable role in this regard [2]. However, GPS is susceptible to various malicious attacks, which pose significant challenges to positioning accuracy as shown in Fig.1. Therefore, it is essential to develop a real-time method for detecting GPS attacks.

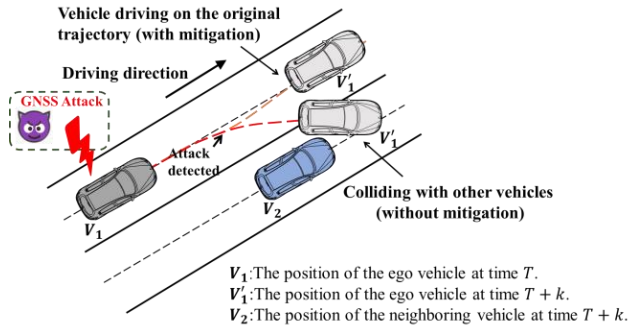

Fig.1: Illustration of GPS attack scenario.

In recent years, GPS attack models have been validated on real vehicles, successfully misleading vehicle trajectories [3].

[1]The authors are within School of Mechanical Engineering, Beijing Institute of Technology, China. Corresponding author: Guoqiang Li (guoqiangli@bit.edu.cn)

This is a serious safety hazard for autonomous vehicles that lack corresponding defense capabilities. Considering the complexity of the GPS system itself, it is unrealistic to avoid GPS attacks from the source. Traditional vehicle information security work mainly focuses on the data security component [4], but this method is powerless against attackers who have knowledge of the internal cryptographic mechanism of the system, and does not consider the interaction between the vehicle and the physical world. At present, with the increasing number of vehicle-mounted sensors, using information cross-validation between different sensors has become a feasible development direction [5]. Utilizing the existing sensors will not only avoid additional costs for the vehicle, but also obtain real-time data [6], creating conditions for real-time GPS attack detection.

Currently, there is a limited body of research on the detection and mitigation of GPS attacks in autonomous vehicles using existing sensor information. In this paper, we propose a novel approach that leverages the integration of cameras and high-precision maps to address this issue. Initially, we align the GPS signal with the high-precision map to extract the lateral distance to the nearest lane. Additionally, we employ a camera-based lane detection algorithm to obtain an independent measurement of the lane's lateral distance. By utilizing these two mutually independent distance measurements, we construct a GPS attack detection framework based on Long Short-Term Memory (LSTM) neural networks. This framework effectively detects anomalies in the temporal data sequence, indicative of potential GPS attacks. After detecting a GPS attack, we propose a mitigation strategy that involves the fusion of data from an in-vehicle Inertial Measurement Unit (IMU), cameras, and the high-precision map using Unscented Kalman filter (UKF). This mitigation strategy aims to preserve the accuracy of the vehicle's positioning by effectively integrating and filtering multi-sensor information.

*Related work*

Real-time protection of vehicles from GPS attacks is a challenging problem in the field of automotive cybersecurity. Reference [7] studies various known attack methods targeting onboard sensors and provides corresponding detection strategies for each type of failure. Reference [8] comprehensively classifies attacks on onboard sensors and identifies two main types of attacks against GPS: GPS interference and GPS spoofing.

GPS attack detection methods mainly include model-based methods and data-driven methods. Model-based methods detect deviations between GPS signal measurements and expected behaviors derived from vehicle mathematical models. Reference [9] proposes a chi-square detector based on residual construction and Kalman filtering for sensor attack detection. The drawback of this method is that the chi-square detector is sensitive to large transient noise, leading to false alarms. To address this issue, [10] and [11] propose using Cumulative Sum (CUSUM) detectors for sensor attack detection. The detector triggers an alert only when the cumulative error between the residual and its expected value exceeds a threshold.

Data-driven methods combine advanced machine learning techniques to train on historical data and learn a set of patterns or rules to determine whether the acquired real-time sensor data is under attack. [12] and [13] compare online data with previously collected data corresponding to specific attacks to detect attack issues. However, this method cannot detect new attacks. Additionally, generating training data is a highly challenging task due to the randomness of network attacks. Since the probability of GPS attacks is small, it is difficult to obtain real abnormal data samples directly from the historical data of moving targets, and it is impossible to obtain all the features of abnormal data directly. Therefore, in this paper, LSTM is chosen as the GPS signal attack detector. The residual values of two lateral distances on different types of roads when GPS is not attacked are used for training. LSTM can learn features from this and use them to detect GPS data abnormalities.

After detecting a GPS attack, mitigating the impact of the attack on vehicle positioning is also crucial. Previous studies have either isolated abnormal data or handed control over to the driver [14], which compromises the stability of vehicle operation and is not suitable for high-level autonomous driving scenarios. In recent research, selecting to park the vehicle in the current lane after detecting a GPS stealth attack [15] partially avoids the greater danger caused by misleading the vehicle into other lanes. However, the compensatory role of other vehicle sensors in position accuracy under GPS failure is not utilized.

In summary, the main contributions of this paper are as follows: (1) a real-time attack detection method integrating image-based lane detection and map-based localization is proposed to identify the GPS attack for autonomous driving. (2) a novel efficient multi-sensor-fusion-based attack mitigation approach is designed to remove the anomaly GPS data and improve the localization accuracy. The method utilizes UKF to fuse the lateral deviation to the lane centerline and the position prediction in the map to provide robust and accurate GPS data. To the best knowledge of the author, this is the first GPS attack mitigation method using image and map based lateral direction localization. (3) Extensive experiments in the Carla simulator and an open-source driving dataset are studied to evaluate the effective and robust performance of the proposed method in detecting and mitigating GPS attack.

The remaining parts of this paper are organized as follows. Section II introduces the method of lateral direction localization using image and map. Section III presents the strategy of GPS attack detection and mitigation. In Section IV, the simulations are reported and conclusions are presented in Section V.

## II. LATERAL DIRECTION LOCALIZATION USING IMAGE AND MAP

In order to achieve the detection and mitigation of GPS attacks as mentioned later in the text, we conducted lateral distance calculations between vehicles and lanes. In the following sections, two independent lateral distance measurements were obtained using a combination of camera-based lane detection and vehicle position estimation matched with lane information from a high-precision map.

### A. Image based lane detection and lateral deviation

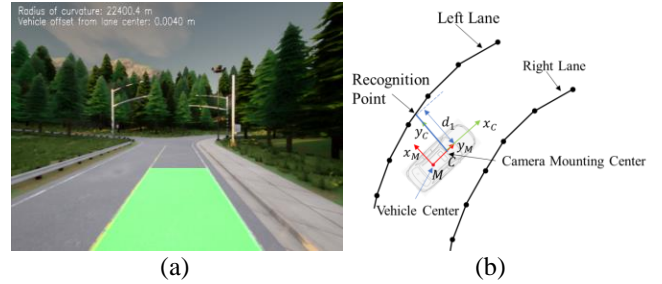

Fig.2: (a) Lane detection (b) Camera-based lateral lane recognition.

Lane detection technology has received significant attention from researchers due to its widespread application in Lane Departure Warning Systems (LDWS) and is now a mature technology. In this study, a classical lane detection technique was adopted, consisting of the following steps: Firstly, image calibration was performed based on the camera's intrinsic and extrinsic parameters, and the original image was preprocessed to locate the region of interest. Then, Gaussian filtering and thresholding were applied to the image for smoothing, resulting in a binary image. A bird's-eye view of the region of interest was obtained through perspective transformation. Subsequently, the sliding window method was used to extract lane markers. Finally, the lane shape was determined by curve fitting based on the center positions of each window. The lane detection result is shown in Fig.2(a).

Lane marking parameter system based on Taylor expansion from a mathematical perspective [17].

$$f(x) = C_0 + C_1 x + C_2 x^2 + C_3 x^3 \qquad (1)$$

The four parameters $C_0$, $C_1$, $C_2$, and $C_3$ represent the lateral distance, slope, curvature, and curvature derivative of the detected lane markings, respectively. Therefore, through this method, we can independently obtain the lateral distance $d_1 = C_0$ between the vehicle and the lane shown in Fig.2(b) using camera sensors.

### B. Map Based Localization

The purpose of the map matching algorithm is to match the estimated position and trajectory of the vehicle obtained from the positioning system with the road lines in a high-definition map, and then project the vehicle's position onto the corresponding lane segment of the high-definition map [18].

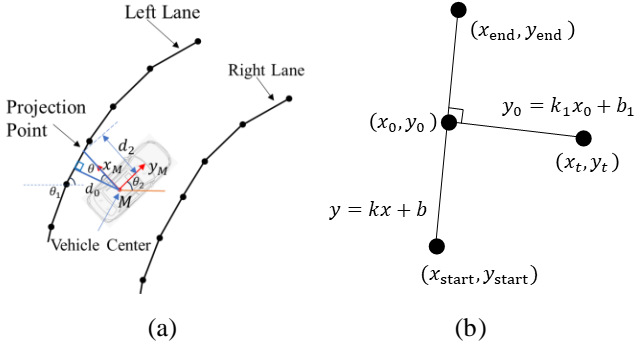

Fig.3: (a) Map-matched lateral lane prediction model. (b) Illustration of vehicle trajectory points matching with a lane segment.

In this study, the specific procedure for obtaining the lateral distance between the vehicle and the lane using map matching is as follows: Firstly, we need to obtain the vehicle's position estimate. Secondly, in conjunction with the high-definition map, the map matching method is employed to find the lane segment that matches the current estimated vehicle position. Finally, the vehicle's position is projected onto the lane, and the projection point coordinates are used to calculate the lateral distance to the lane.

High-precision maps can provide data such as lane types, lane IDs, and lane sampling point coordinates. By searching for lane information within a certain range near the vehicle's position, we can obtain a series of lane segments that meet the requirements for vehicle driving. As shown in Fig.3(b), the starting point $(x_{start}, y_{start})$ and endpoint $(x_{end}, y_{end})$ are connected to form a lane segment. For each segment the projection point coordinates $(x_0, y_0)$ are obtained by projecting the location estimate coordinates $(x_t, y_t)$ onto each lane segment:

$$kx_0 + b = k_1 x_0 + b_1 \quad (2)$$

$$y_0 = kx_0 + b \quad (3)$$

where $k$ and $b$ are the slope and intercept of each lane segment. $k_1$ and $b_1$ are the slope and intercept of the perpendicular lines. Then, we can determine whether the matching is successful by judging whether the horizontal coordinate of the projected point falls within the horizontal coordinates of the start and end points of the lane section.

$$x_0 \geq \min\{x_{start}, x_{end}\} \text{ and } x_0 \leq \max\{x_{start}, x_{end}\} \quad (4)$$

In order to determine the lateral distance between the vehicle and the lane line, we need to calculate the closest distance $d_0$ between location estimate and matched lane segments.

$$d_0 = \min\left\{\left|\frac{-k_k x_t + y_t - b_k}{\sqrt{k_k^2 + 1}}\right|\right\} \quad (5)$$

where $k_k$ and $b_k$ represent the slope and intercept of the $k$-th lane segment that satisfies the requirement. In addition, the vehicle's heading angle $\theta_2$ can be obtained from the IMU. As shown in Fig.3(a), the lateral distance $d_1$ can be calculated as follows:

$$d_1 = \frac{d_0}{\cos\theta} = \frac{d_0}{\cos\theta'} = \frac{d_0}{\cos(\theta_1 - \theta_2)} \quad (6)$$

## III. GPS ATTACK DETECTION AND MITIGATION

In order to achieve detection and mitigation of GPS, during the data acquisition phase, as shown in Fig.4., we obtained vehicle position information provided by GPS, vehicle speed and angular velocity information provided by the wheel speed sensor and gyroscope in the IMU, RGB images provided by the camera, and lane semantic information provided by a high-precision map. In the detection phase, we performed lane detection on the camera images and map matching with GPS data to obtain two lateral distances, which were then used to calculate the residuals and inputted into an LSTM detector for detection. When the detection result is normal, the IMU and GPS data are fused using an Extended Kalman Filter (EKF). When the detection result is abnormal, the GPS data is isolated, and the IMU data is fused with the camera and HD map data using a UKF to mitigate the estimation error of the vehicle's position.

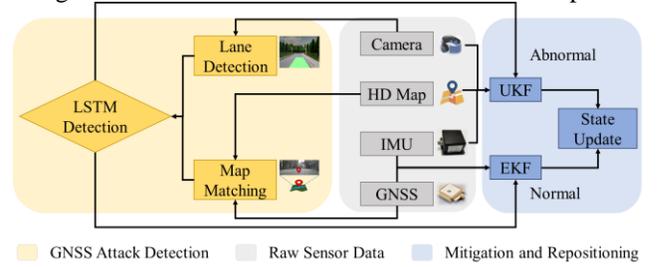

Fig.4: The framework of GPS detection and mitigation

### A. Attack Models

GPS signals may be subject to various types of attacks during actual use, mainly including interference, replay, and spoofing attacks. Among them, spoofing attacks targeting GPS are the most widespread and difficult to identify. The principle is to inject false data into normal GPS signals. It gradually degrades sensor readings over time, thus introducing undetected errors into state predictions.

This article mainly considers two typical GPS spoofing attack models, namely the constant bias attack and the stealth attack. Other types of spoofing attack signals can be approximated using these two attacks.

The constant bias attack involves adding a constant bias to the correct GPS signal for a period of time, causing the GPS reading to temporarily deviate from the true value. In actual attacks, attackers can launch deviation attacks to mislead vehicles by adding lateral and vertical offsets (or both) to real GPS readings.

$$\widetilde{y}_{k,gps} = y_{k,gps} + c \quad (k \in [t_\zeta, t_\phi]) \quad (7)$$

$\widetilde{y}_{k,gps}$ is the GPS signal under constant bias attack. $c$ is the stable bias vector added to the normal signal. Under continuous deviation attacks, vehicles may mistakenly believe that they are in the wrong lane position, leading to unreasonable actions.

The stealth attack involves injecting a series of incrementally increasing biases into the real measurements,

causing the aircraft to gradually deviate from the true trajectory. In mathematical terms, the received GPS measurement can be represented as:

$$\widetilde{\boldsymbol{y}}_{k,gps} = \boldsymbol{y}_{k,gps} + \boldsymbol{\varphi}_k\bigl(k \in [t_\zeta, t_\phi]\bigr) \quad (8)$$

where $\boldsymbol{\varphi}_k$ is an attack signal that is carefully designed to avoid triggering attack detectors, making stealth attacks more deceptive than constant bias attacks. For example, noise attacks can be achieved by representing $\boldsymbol{\varphi}_k$ as adding data from a distribution with a large variance. In this article, we use an advanced stealth spoofing strategy [19] to construct the attack signal in exponential form. As shown in the formula (9), it can be seen that in discrete time, the bias gradually increases from small undetectable values, and its change follows:

$$\boldsymbol{\varphi}_k = \gamma * \delta^k \quad (9)$$

where γ and δ are used to control the speed of change of the attack. Because at the beginning, the GPS signal under attack cannot be detected, but instead is fused and destroys the data fusion framework.

### B. LSTM-based attack data detection method

In this paper, a widely used artificial neural network called LSTM is employed as the attack detector. During non-attack periods, LSTM learns the residuals of lateral distances between the vehicle and the lane under different road conditions to predict future values based on the learned features from the input data during the detection phase.

In the LSTM-based attack detection method, the mean squared error (MSE) $\xi$ is used as the loss function, which is the sum of the squared differences between the predicted values and the target value. This value is used as a criterion for measuring whether the GPS signal is under attack.

$$\xi = \frac{1}{n} * \sum (z_i - \hat{z}_i)^2 \quad (10)$$

where $n$ represents the number of samples, $z_i$ represents the actual residuals, and $\hat{z}_i$ represents the predict residuals. We set $\gamma$ as MSE threshold. If $\xi > \gamma$, it means GPS is under attacked; on the contrary $\xi \leq \gamma$, it shows GPS data is safe. When $\gamma$ is set appropriately, it can help balance precision and recall. A smaller threshold can reduce detection lag and detect more attacked GPS data. However, this can also lead to increased false alarm rates and decreased accuracy. On the other hand, a larger threshold may prolong detection time or completely miss alarms.

### C. GPS mitigation with UKF

In the mitigation part of this paper, IMU data is fused with camera and high-precision map information. The lateral distance of lane detection from the camera at each moment is taken as the observation $\boldsymbol{y}_k$. Based on the predicted position value $\breve{\boldsymbol{x}}_k$ at each moment, the predicted lateral distance $\widehat{\boldsymbol{y}}_k$ is obtained through unscented propagation and map matching. The two are fused to update the vehicle's position $\widehat{\boldsymbol{x}}_k$ and the covariance matrix $\widehat{\boldsymbol{P}}_k$ at each moment. Since this is a strongly nonlinear system, adopting UKF can achieve better localization performance compared to EKF. The computation procedure is as follows.

In the prediction stage, to propagate the vehicle state from time $(k-1)$ to time $k$, the Unscented Transform is applied using the current best guess of the mean and covariance. For an *N*-dimensional probability density function (PDF), $\mathcal{N}(\mu_x, \Sigma_{xx})$, we need $2N+1$ sigma points. In this research, considering that the vehicle's position estimation is in a two-dimensional space, we select $N = 2$.

$$\breve{\boldsymbol{x}}_k^{(i)} = \boldsymbol{f}_{k-1}\bigl(\widehat{\boldsymbol{x}}_{k-1}^{(i)}, \boldsymbol{u}_{k-1}, \boldsymbol{0}\bigr) \; i = 0 \dots 2N \quad (11)$$

where $\widehat{\boldsymbol{x}}_{k-1}^{(i)}$ means the sigma points of time $(k-1)$. $\boldsymbol{u}_{k-1}$ is the control variable of time $(k-1)$. Then the predicted mean $\breve{\boldsymbol{x}}_k$ covariance matrix $\breve{\boldsymbol{P}}_k$ can be calculated as follows.

$$\breve{\boldsymbol{x}}_k = \sum_{i=0}^{2N} \alpha^{(i)} \breve{\boldsymbol{x}}_k^{(i)} \quad (12)$$

$$\breve{\boldsymbol{P}}_k = \sum_{i=0}^{2N} \alpha^{(i)} \bigl(\breve{\boldsymbol{x}}_k^{(i)} - \breve{\boldsymbol{x}}_k\bigr)\bigl(\breve{\boldsymbol{x}}_k^{(i)} - \breve{\boldsymbol{x}}_k\bigr)^T + \boldsymbol{Q}_{k-1} \quad (13)$$

where $\boldsymbol{Q}_{k-1}$ is the additive process noise, $\alpha^{(i)}$ is the weight.

In the update step, we need to obtain the Kalman gain.

$$\boldsymbol{K}_k = \boldsymbol{P}_{xy} \boldsymbol{P}_y^{-1} \quad (14)$$

where $\boldsymbol{P}_y$ is the covariance of predicted measurements and $\boldsymbol{P}_{xy}$ is the cross-covariance. Then we can calculate the corrected mean and covariance as follows.

$$\widehat{\boldsymbol{x}}_k = \breve{\boldsymbol{x}}_k + \boldsymbol{K}_k(\boldsymbol{y}_k - \widehat{\boldsymbol{y}}_k) \quad (15)$$

$$\widehat{\boldsymbol{P}}_k = \breve{\boldsymbol{P}}_k - \boldsymbol{K}_k \boldsymbol{P}_y \boldsymbol{K}_k^T \quad (16)$$

By adopting the UKF to fuse sensor data, we alleviate the impact of GPS being compromised and unable to provide accurate position information, enabling the vehicle to maintain a high level of localization accuracy.

## IV. EXPERIMENT ANALYSIS

To validate the accuracy and real-time performance of the proposed GPS attack detection and mitigation methods, we conducted tests on Driving Scenario Designer (DSD), Carla, and the publicly available real-world dataset nuScenes. DSD provides an idealized simulated scenario, Carla offers a simulation environment closely resembling real vehicle driving, and nuScenes dataset is a widely used real-world dataset. Transitioning from virtual environments to real-world scenarios, our goal was to verify the feasibility of applying the proposed attack detection and mitigation strategies to future autonomous vehicles.

On each platform, we collected GPS, IMU, camera, and lane semantic information near the vehicle trajectory. To simulate attacks, we injected designed GPS attack models into the raw GPS data.

In the detection phase, we first trained an LSTM detector using lateral distance residuals obtained from driving on different trajectories on each platform without any attacks. Based on this, we calculated the alarm accuracy and detection time of our proposed framework. Additionally, during the alarm process, we applied the proposed mitigation measures to correct the vehicle's positioning. Finally, we computed the average maximum RMSE of the vehicle during the GPS attack stage, both with and without the mitigation measures.

In the following section, we will discuss the performance of the proposed framework.

*A. Attack Design*

To simulate GPS attacks, we randomly selected the start time and duration of the attack. The attack modified the real GPS data, and the modified data was passed to the vehicle for position estimation.

We compared the LSTM-based attack detection method with baseline methods, including cumulative sum (CUSUM) detector [20] and Isolation Forest (IF) detector [21]. We evaluated the performance of the method using multiple commonly used metrics, including F1 score, accuracy, recall, and detection lag. The first three metrics evaluate the accuracy of the detection, ranging from 0 to 1. Detection lag measures the time delay between injecting the attack signal and the detector detecting the attack. If the GPS measurement is tampered with but not detected in a timely manner, the affected vehicle will absorb this information for position estimation, leading to trajectory deviation.

TABLE I. DETECTION PERFORMANCE OF THE PROPOSED METHOD AND BASELINE METHOD UNDER TWO TYPES OF ATTACKS

| Attack Scenario | CUSUM | | | |
|---|---|---|---|---|
| | *Precision* | *Recall* | *F1* | *Delay(s)* |
| Constant Bias | 0.82 | 0.95 | 0.88 | 0.2 |
| Spoofing | 0.73 | 0.85 | 0.79 | 0.9 |
| Attack Scenario | IF | | | |
| | *Precision* | *Recall* | *F1* | *Delay(s)* |
| Constant Bias | 0.93 | 0.95 | 0.94 | 0.2 |
| Spoofing | 0.94 | 0.87 | 0.91 | 0.8 |
| Attack Scenario | LSTM | | | |
| | *Precision* | *Recall* | *F1* | *Delay(s)* |
| Constant Bias | 0.98 | 1 | **0.99** | **0** |
| Spoofing | 0.98 | 0.9 | **0.94** | **0.6** |

*B. Attack Detection*

In this experiment, we first applied continuous bias attacks to the normal GPS signal, and the detection results are shown in Fig.5(a). Table I provides the average results of multiple experiments. The results show that all methods were able to detect the start of the attack within 0.2s. However, the LSTM-based detector achieved an F1 score of 0.99, significantly higher than the CUSUM detector and slightly higher than the Isolation Forest detector. This is because LSTM can effectively differentiate between normal and abnormal data through feature learning.

The results of stealthy attack detection are shown in Fig.5(b). According to the average results in Table I, the LSTM-based method achieved an F1 score of 0.94, higher than the other two detectors. Under stealthy attacks, the CUSUM detector took 0.9s to trigger an alarm. Meanwhile, the Isolation Forest detection method is sensitive to anomalous signals, generating multiple false alarms due to significant noise in normal GPS measurements before the attack occurs.

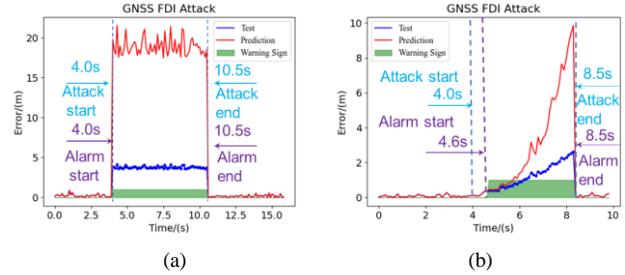

Fig.5: (a) Detection results under GPS constant bias attack. (b) Detection results under GPS stealthy attack

*C. Attack mitigation*

After detecting an attack, we apply the previously mentioned UKF method to reduce the vehicle's position offset. Under GPS attack, we obtain the estimated vehicle position with and without the mitigation measure. We calculate the RMSE in the x, y directions and overall, between the estimated vehicle position and the actual position.

$$RMSE = \sqrt{\frac{\sum_{i=1}^{n}(X_{est}-X_{act})^2}{n}} \quad (17)$$

where $X_{est}$ is the estimated position, $X_{act}$ is the actual position, $n$ is the number of dimensions. The following sections provide a detailed analysis based on different experimental scenarios.

*D. Evaluation in DSD Simulator*

In the DSD experiments, we configured high-precision sensors and denser lane sampling points (1 meter) to explore the positioning accuracy that can be achieved under ideal conditions using the proposed mitigation method. We conducted simulations on straight roads and the curved road shown in Fig.6. From the results in Table II, it can be observed that the overall RMSE value of the vehicle under the mitigation method is 0.032m, achieving centimeter-level positioning accuracy.

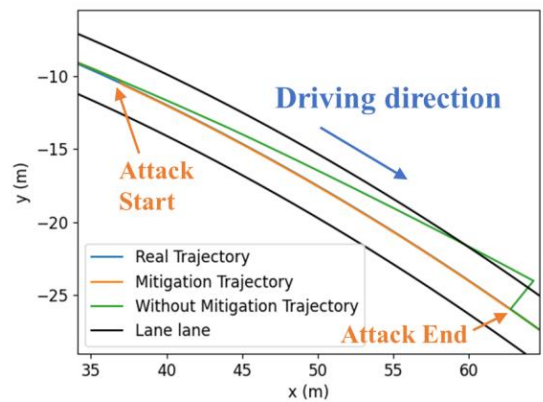

Fig.6. Comparison of positioning performance when the GPS is attacked: slowly turning driving. Experimental environment: DSD.

TABLE II. RMSE VALUES IN DIFFERENT SCENARIOS UNDER GPS ATTACKS. EXPERIMENTAL ENVIRONMENT: DSD.

| RMSE(m) | $x$ direction | $y$ direction | $(x, y)$ |
|---|---|---|---|
| Mitigation | 0.012 | 0.025 | 0.032 |
| Without Mitigation | 1.52 | 1.54 | 2.16 |

### E. Evaluation in Carla Simulator

To make the experiments more realistic, we used sensors with larger errors in the Carla simulation and conducted attacks and mitigations in more challenging scenarios, such as the curved road section shown in the Fig.7. From the simulation results in Table III, it can be observed that when GPS attacks cause the estimated vehicle position to deviate by more than 4.7 meters, the average positioning accuracy of the vehicle remains within 0.4 meters through the adoption of the mitigation measure. Considering that the positioning error mainly comes from the longitudinal direction of the vehicle, the vehicle can still drive smoothly within its own lane on sharp turns.

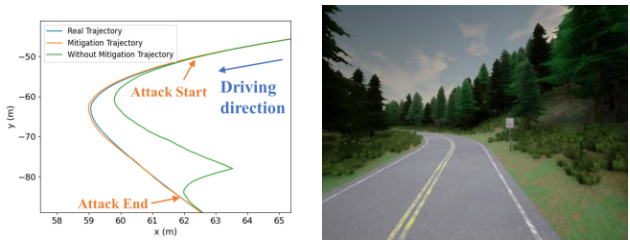

Fig.7: Comparison of positioning performance when the GPS is attacked: sharply turning driving. Experimental environment: Carla.

TABLE III. RMSE VALUES IN DIFFERENT SCENARIOS UNDER GPS ATTACKS. EXPERIMENTAL ENVIRONMENT: CARLA.

| RMSE(m) | $x$ direction | $y$ direction | $(x, y)$ |
|---|---|---|---|
| Mitigation | 0.240 | 0.170 | 0.358 |
| Without Mitigation | 2.410 | 2.380 | 4.789 |

### F. Evaluation in nuScenes Dataset

Compared with the lane sampling point coordinates output in the simulated scenario, the semantic map sampling points provided in the dataset are sparser. We selected different driving segments from the dataset for validation, one of which is shown in Fig.8. From the simulation results in Table IV, it can be observed that under the adoption of the mitigation measure, the average positioning accuracy of the vehicle is within 0.4 meters. This indicates that even when using real-world vehicle datasets, our proposed method can still ensure safe navigation for vehicles within their lanes under GPS attack scenarios.

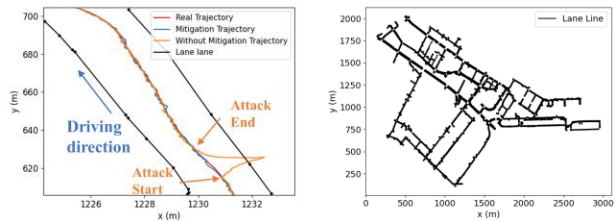

Fig.8: Comparison of positioning performance when the GPS is attacked: slowly turning driving. Experimental environment: nuScenes.

TABLE IV. RMSE VALUES IN DIFFERENT SCENARIOS UNDER GPS ATTACKS. EXPERIMENTAL ENVIRONMENT: NUSCENES.

| RMSE(m) | $x$ direction | $y$ direction | $(x, y)$ |
|---|---|---|---|
| Mitigation | 0.316 | 0.137 | 0.312 |
| Without Mitigation | 2.062 | 2.049 | 2.908 |

## V. CONCLUSION

This paper has presented a novel anomaly detection and mitigation method against GPS attacks that utilizes onboard cameras and high-precision maps is proposed to ensure accurate vehicle localization. This method first obtains the lateral distance information between the vehicle and the lane from two independent data sources: lane detection and map matching. Based on this information, a GPS attack detection method based on LSTM is constructed. When an attack is detected, a multi-source fusion positioning method using a UKF is proposed to counter GPS attacks and improve positioning accuracy. Experimental results on simulation software DSD, Carla, and real-world dataset nuScenes demonstrate that the proposed method outperforms existing methods in terms of real-time GPS attack detection and effective mitigation.